\pdfoutput=1

\documentclass[11pt]{article}

\usepackage[final]{acl}

\usepackage{times}
\usepackage{latexsym}

\usepackage[T1]{fontenc}

\usepackage[utf8]{inputenc}

\usepackage{microtype}

\usepackage{inconsolata}

\usepackage{mdframed}
\usepackage{graphicx}
\usepackage{float}
\usepackage{xurl}

\usepackage{amssymb}
\usepackage{amsmath}
\usepackage{epsfig,subfigure,caption}
\usepackage{algpseudocode}
\usepackage[normalem]{ulem}
\usepackage[linesnumbered,algoruled,boxed,noend]{algorithm2e}
\usepackage{listings}  
\lstdefinestyle{promptcode}{
    basicstyle=\footnotesize\ttfamily,
    breaklines=true,
    breakatwhitespace=true,
    columns=fullflexible,
    keepspaces=true,
    showstringspaces=false,
    frame=none,
    xleftmargin=0pt,
    aboveskip=2pt,
    belowskip=2pt,
    literate={_}{\_}{1}
}
\usepackage{xcolor}     
\usepackage{color, colortbl}
\usepackage{multirow,booktabs,hhline,tabularx}
\usepackage{bm}
\usepackage[linesnumbered,algoruled,boxed,noend]{algorithm2e}
\usepackage{wrapfig}
\usepackage{verbatimbox}
\usepackage{kantlipsum}
\usepackage{fancyvrb}
\usepackage{fvextra}
\usepackage{xcolor}
\usepackage[many]{tcolorbox}
\usepackage{needspace}
\usepackage[htt]{hyphenat}
\usepackage{enumitem}
\setlist[itemize]{noitemsep, topsep=2pt, parsep=0pt, partopsep=0pt}
\usepackage{pifont}

\tcbset{
    myboxstyle/.style={
        colback=yellow!10!white, 
        colframe=yellow!50!black, 
        fonttitle=\bfseries,
        coltitle=black,
        boxrule=0.75mm,
        boxsep=1pt,
        width=\linewidth,
        sharp corners,
        leftrule=1mm,
        left=5pt,
        right=5pt,
        top=5pt,
        bottom=4pt,
        fontupper=\small,
        fontlower=\small,
        noparskip,
        compress page=all,
        before skip=6pt,
        after skip=4pt,
        lines before break=8,
        pad at break=1mm,
        before={\par\needspace{12\baselineskip}\noindent},
        before upper={\setlength{\parskip}{0pt}}
    }
}
\newcommand{\secref}[1]{\S\ref{#1}}

\newcommand{\ourdata}{\textsc{Textbook-Exam}\xspace} 
\newcommand{\ours}{\textsc{Quest}\xspace}
\definecolor{darkgreen}{RGB}{95,129,63}
\definecolor{darkorange}{RGB}{184, 96, 41}

\title{Which Questions Improve Learning the Most? \\ Utility Estimation of Questions with LM-based Simulations}

\author{
    Dong-Ho Lee\thanks{Authors contributed equally},~
    Hyundong Cho\footnotemark[1],~
    Jonathan May\thanks{Equal advising contribution},~
    Jay Pujara\footnotemark[2] \\
    Information Sciences Institute, University of Southern California \\
     {\small 
        \texttt{dongho.lee@usc.edu},~
        \texttt{\{jcho, jonmay, jpujara\}@isi.edu}
    }\\
}

\begin{document}
\maketitle
\begin{abstract}

Asking questions is fundamental for deeper understanding, yet evaluating and generating questions that improve downstream outcomes remains a challenging problem.
To address this gap, we introduce \ours (\textbf{Q}uestion \textbf{U}tility \textbf{E}stimation with \textbf{S}imulated \textbf{T}ests), a framework that offers a new outcome-driven paradigm for question evaluation and generation, instead of indirect metrics that are based on the underlying content of the question and its answer. 
\ours uses language models to simulate learners that ask questions while studying a textbook chapter and quantifies the utility of a question as its direct contribution to understanding as measured by the learner's exam performance.
To analyze the effectiveness of \ours, we curate \ourdata, a benchmark that aligns textbook sections with exam questions across five academic disciplines.
Using \ours, we filter for high-utility questions and fine-tune question generators via rejection sampling and show that questions generated by \ours-trained models improve simulated exam scores by 18.5\% compared to the next strongest baseline.
Furthermore, we demonstrate the practical real-world value of \ours by showing that simulation-computed utility of a question is moderately positively correlated with its usefulness as perceived by human evaluators.

\end{abstract}

\section{Introduction}

\begin{figure}[t!]
    \centering
    \begin{minipage}{\columnwidth}
    \centering
    \includegraphics[width=\columnwidth]{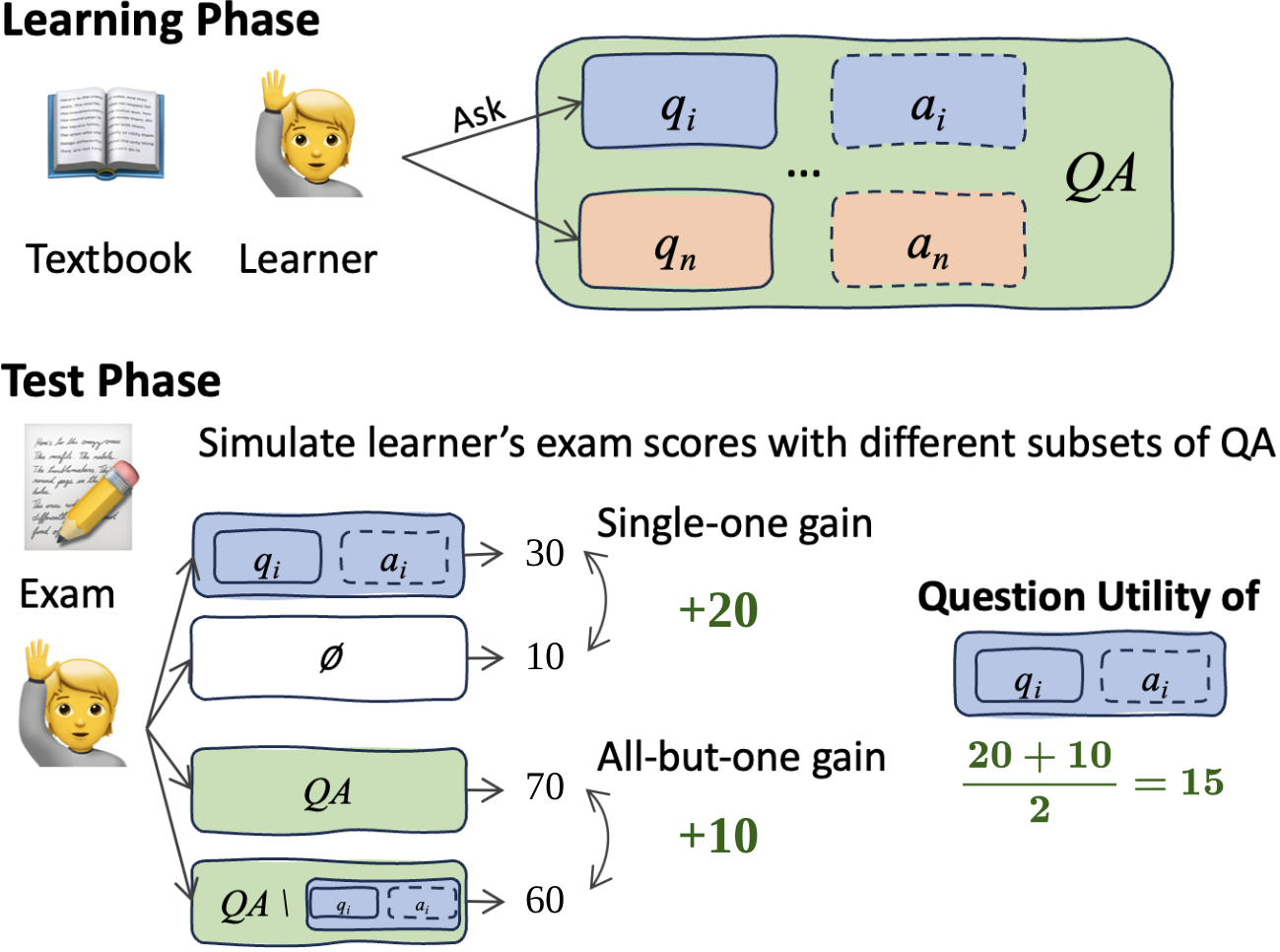}
    \caption{\textbf{Question utility} is defined as the direct impact of a question-answer (QA) pair on a learner's understanding, measured by simulated exam performance improvement. \ours estimates this marginal benefit by comparing simulated learners with and without access to the QA pair, enabling question evaluation and generation based on direct outcome rather than indirect metrics.
    }
    \vspace{-0.5cm}
    \label{fig:intro}
\end{minipage}
\end{figure}


Asking questions is a crucial skill for improving understanding.
Effective teachers pose questions to stimulate critical thinking~\cite{bloom1956handbook, tofade2013best}, while high-performing students actively ask clarifying questions~\cite{abbott1980teaching, brassell2008comprehension, davoudi2015systematic}. 
But among the many questions one could ask, how can we identify those that are beneficial for improving downstream learning outcomes, and how can we systematically generate such questions?


The psychology literature offers general heuristics~\cite{graesser2010good, vale2013value} while recent NLP works have focused on quantifying the quality of \textit{inquisitive questions}, which are learner-generated and curiosity-driven queries~\cite{scialom-staiano-2020-ask}, using indirect metrics such as salience~\cite{wu2024questions} or expected information gain~\cite{rao-daume-iii-2018-learning, keh-etal-2024-asking}.
Although such proxies reflect properties like relevance or informativeness, they do not measure a question's impact on downstream outcomes.


We propose a new direction: measuring a question's utility by its \textit{direct} impact on downstream outcomes.
Inspired by recent work leveraging language models (LMs) as simulators of human behavior~\cite{zelikman-etal-2023-generating, park2024generative, zhang2024simulating, he2024evaluating}, we propose \ours (\textbf{Q}uestion \textbf{U}tility \textbf{E}stimation with \textbf{S}imulated \textbf{T}ests), a framework that estimates question utility -- how much a question helps a learner by simulating its contribution to a downstream task -- and uses high-utility questions to fine-tune question generators.


In \ours, as shown in \autoref{fig:intro}, an LM role-plays as a novice learner who poses questions while studying learning materials and receives answers from another LM that acts as a teacher (\secref{ssec:quest-question-generation}).
We estimate the \textit{utility} of each question by simulating exam performance taken with different QA subsets, without access to the original text, to isolate their marginal benefit (\secref{ssec:quest-evaluation}).
\ours can then fine-tune question generators via rejection sampling~\cite{bai2022constitutional} with high-utility questions that surpass a certain threshold (\secref{ssec:quest-train}).

To evaluate \ours, we curate \ourdata, a dataset in which each sample consists of a chapter with multiple sections and corresponding exam questions (\secref{sec:textbook-exam}).
Using \ourdata, we experiment with various question generation strategies and quality metrics.
We find that \ours-trained question generators produce questions that lead to an average improvement of 18.5\% in simulated exam scores across five academic subjects (\secref{ssec:overall-performance}).
In addition, correlation analysis reveals that indirect metrics such as salience and expected information gain are not correlated with utility and that optimizing for them yields suboptimal results (\secref{ssec:evaluation-metrics}).
While exam question similarity is weakly correlated with utility, models fine-tuned to predict exam questions from the chapter content produce low-utility questions.
Lastly, through a human evaluation study with a setup similar to \ours, we find that simulation-computed utility is moderately correlated with the usefulness of the question perceived by human evaluators, thus validating the practical value of \ours (\secref{ssec:human-evaluation}).

In summary, our main contributions are: (\textit{i}) \ours, a framework that quantifies the utility of a question by estimating its direct impact on downstream outcomes through LM-based simulations of a learner;
(\textit{ii}) \ourdata, a benchmark that aligns textbook sections with exam questions across five academic disciplines, to evaluate \ours;
(\textit{iii}) demonstrating that \ours-trained question generators produce questions that result in significant improvements in simulated exam scores on \ourdata;
(\textit{iv}) establishing the practical and unique value of \ours by showing that simulation-computed utility of a question is moderately correlated with the usefulness of the question perceived by human evaluators but not correlated with indirect metrics.

\section{\ours}
\label{sec:framework}

In this section, we present \ours,\footnote{All our code and data are available at \url{https://github.com/usc-isi-i2/quest}.} a framework for measuring a question's utility based on its impact on downstream learning outcomes and fine-tuning question generators with high-utility questions. 
We first formulate the problem and define our core evaluation metrics.
Then, we provide an overview of the \ours framework and describe each component in detail.

\subsection{Problem Formulation}

Our objective is to generate questions that effectively support learners studying a given document $D$.
We assume access to a document $D$ consisting of sequential sections $S_1, \dots, S_k$, and a set of exam questions $E = \{e_1, \dots, e_m\}$ designed to evaluate understanding of the content in $D$.

Given this setup, we develop a question generator $M_q$ that produces a set of questions $Q = \{q_1, \dots, q_n\}$.
Each question $q_i$ is paired with an answer $a_i$, generated independently using a fixed answer generator $M_a$ based on general parametric knowledge.
We assume that all answers are approximately correct and consistent across questions, so that any observed difference in learner performance can be attributed to the quality of the question itself.
The resulting set of question–answer pairs $\text{QA} = \{(q_1, a_1), \dots, (q_n, a_n)\}$ is used to support a simulated learner, who attempts the exam $E$ having access only to these QA pairs.
We exclude the document 
$D$ to isolate the instructional value of the questions, avoiding confounding effects from the document content.
The learner's performance on the exam serves as a proxy for the instructional quality of the generated questions.

To quantify this, we introduce two evaluation metrics:
(1) \textit{exam score} is defined as the performance of the simulated learner $M_l$ on $E$, using only the QA pairs as study material.
This score, measured as average accuracy across exam questions, ranges from 0 to 100 and reflects how well the questions prepare the learner;
(2) \textit{utility} $u_i$ of each question $q_i$ is defined  as its marginal contribution to the \textit{exam score}.
Unlike traditional educational frameworks such as Item Response Theory (IRT), which model latent traits such as learner ability or test item difficulty~\citep{harvey1999item, sumita-etal-2005-measuring, lalor-etal-2016-building, zhou-etal-2019-intelligent}, our formulation of utility focuses on the \textit{interventional} effect of a question asked while learning on the outcome (i.e. exam performance) for a fixed simulated learner with no prior knowledge of $D$ who learns only from the provided \text{QA} pairs.

\subsection{Overview}
\label{ssec:quest-overview}

\begin{figure}[t!]
    \centering
    \begin{minipage}{\columnwidth}
    \centering
    \includegraphics[width=\columnwidth]{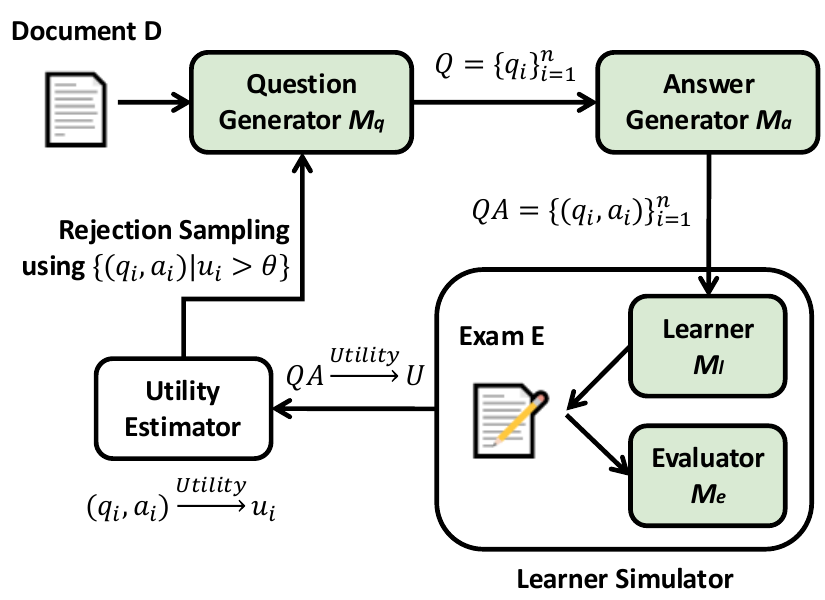}
\end{minipage}
\caption{\textbf{\ours Framework.} $M_q$ generates questions and $M_a$ provides answers. A simulated learner $M_l$ studies the QA pairs and takes an exam $E$, which is scored by evaluator $M_e$. The utility estimator measures each question’s impact on exam performance, and only high-utility questions are retained to refine $M_q$.}
    \label{fig:framework}
\end{figure}

\ours consists of the following four components, corresponding directly to the outcome-based formulation above:
(1) \textbf{question generator} ($M_q$) takes a document $D$ and generates a set of questions $Q = \{q_1, \dots, q_n\}$, grounded in the document's content;
(2) \textbf{answer generator} ($M_a$) independently produces an answer $a_i$ for each question $q_i$ using general parametric knowledge, forming the set of QA pairs $\text{QA} = \{(q_1, a_1), \dots, (q_n, a_n)\}$;
(3) \textbf{learner simulator}, composed of a learner model $M_l$ and an evaluator $M_e$, assesses the effectiveness of QA by having $M_l$ attempt an exam $E$ using only the QA pairs, and $M_e$ compute the resulting exam score;
(4) \textbf{utility estimator} runs the learner simulator under different QA subsets to estimate each pair's marginal contribution to the learner’s exam performance. This enables selection of high-utility questions for refining $M_q$ via rejection sampling.
See Figure~\ref{fig:framework} for a visual summary. Additional details, such as the prompts we use and validation of the reliability of answers by $M_a$ and scores produced by $M_e$, are provided in Appendix \ref{appendix:quest-details}.

\subsection{Generating Questions}
\label{ssec:quest-question-generation}

The \textbf{Question Generator} (\(M_q\)) produces a set of questions \(Q\) from the input document \(D\).
For each section \(S_k\), which represents the part of the document the learner is currently reading (the \textit{anchor}), it considers both \(S_k\) and its preceding context \(C_k = \{S_1, \dots, S_{k-1}\}\) to generate a set of questions \(Q_k = M_q(S_k, C_k)\).

This approach ensures that the questions are informed by both local and global context, following prior work on inquisitive question generation~\cite{wu2024questions}.
Each question \(q_i\) is then paired with an answer \(a_i = M_a(q_i)\), generated independently by the \textbf{Answer Generator} (\(M_a\)) using general parametric knowledge.
The resulting QA pairs \(\text{QA} = \{(q_1, a_1), \dots, (q_n, a_n)\}\) serve as input for learner simulation and utility estimation.

\subsection{Evaluating Question Generators via \textit{Exam Score}}
\label{ssec:quest-evaluation}
To isolate the effect of using different question generators $M_q$, the learner’s exam performance is simulated with access to only the generated QA pairs; i.e.
$M_l$ answers the exam $E$ based solely on $\text{QA} = \{(q_1, a_1), \dots, (q_n, a_n)\}$, producing responses $P_{QA} = M_l(E, \text{QA})$.


These responses are then scored by an evaluator model $M_e$, which either compares them to ground-truth answers or uses its own parametric knowledge when ground truth is unavailable or insufficient, such as for open-ended or free-form questions.
This yields an exam score $s_{QA} = M_e(E, P_{QA})$, which serves as a direct measure of how effectively the generated questions support learning.

\subsection{Improving Question Generator via \textit{Utility}}
\label{ssec:quest-train}
Our objective is to improve the question generator $M_q$ by selecting and learning from high-utility questions, those that most effectively enhance exam performance when studied by the learner simulator.

To do this, we estimate the utility $u_i$ of each QA pair $(q_i, a_i)$ based on its marginal contribution to exam performance. 
The utility estimator runs the learner simulator under two perturbations:
(1) the \textit{single-one gain}, defined as the exam score when only $(q_i, a_i)$ is provided: $s_{\{(q_i, a_i)\}} - s_{\emptyset}$, which captures how much a learner can benefit from the pair in isolation; and
(2) the \textit{all-but-one gain}, defined as the score drop when $(q_i, a_i)$ is removed from the full QA set: $s_{QA} - s_{QA \setminus \{(q_i, a_i)\}}$, which captures how essential the pair is in context.
While $s_{\emptyset}$ ideally reflects a zero-knowledge baseline, it may be non-zero in practice due to the learner's parametric knowledge (e.g., general world knowledge or memorized patterns from pretraining). 
We treat this as a background prior and compute utility relative to this baseline.
The final utility is computed as the weighted average, with $\alpha$ as a hyperparameter, of the two gains:

{\small
\begin{equation*}
u_i = \alpha\big(s_{\{(q_i,a_i)\}} - s_{\emptyset}\big) + (1-\alpha)\big(s_{QA} - s_{QA \setminus \{(q_i,a_i)\}}\big)
\end{equation*}
}
We retain only the QA pairs with $u_i \geq \theta$, and update $M_q$ using these high-utility examples via a rejection sampling strategy~\cite{bai2022constitutional}. We set $\alpha=0.5$ unless mentioned otherwise. 

\section{\ourdata: A Dataset for Outcome-Based Question Evaluation}
\label{sec:textbook-exam}

\begin{figure}[t]
    \centering
    \includegraphics[width=\linewidth]{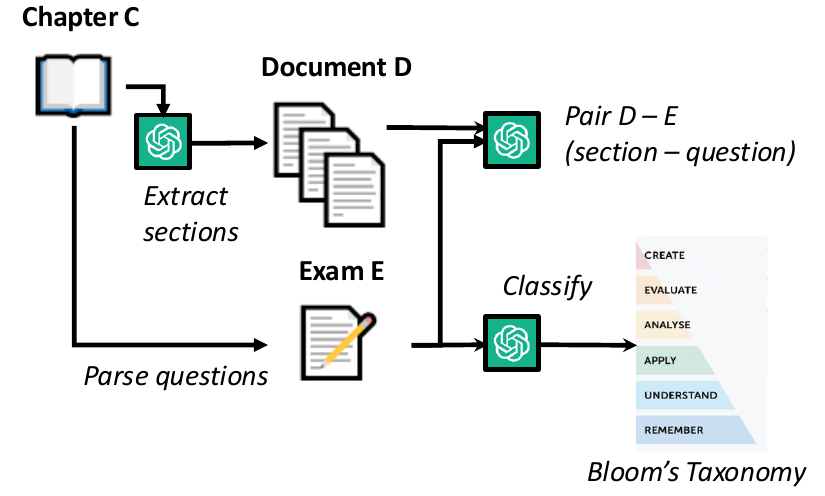}    
    \caption{\textbf{\ourdata Curation Pipeline.} Given a chapter \(C\), we segment its content into document sections \(D = \{S_1, \dots, S_k\}\) and extract exam questions to form the exam \(E\). Each exam question is classified by Bloom’s taxonomy and aligned to relevant sections.}
    \label{fig:dataset-overview}
    \vspace{-0.5cm}
\end{figure}

To evaluate the effectiveness of \ours in diverse learning settings, we construct \ourdata, a dataset composed of documents $D$ and corresponding exam questions $E$, derived from real-world textbooks. 
Each entry is designed to simulate a realistic learning task where a learner studies $D$ and is later evaluated on $E$.

\subsection{Data Processing}
\label{ssec:textbook-exam-pipeline}

We start with college-level textbooks from the OpenStax repository.\footnote{\url{https://github.com/philschatz/textbooks}}  
Each textbook is divided into chapters, and for each chapter \(C\), we extract the main content to form the document \(D\) and the end-of-chapter exam questions to form the exam \(E\).


\paragraph{Sectioning the Document.}
To simulate a learner progressing incrementally through a document, we divide each \(D\) into sections \(\{S_1, \dots, S_k\}\) using an LM-based segmentation method. 
We provide few-shot examples of sections for each subject for consistent structuring across subjects. 
This setup allows for evaluating how well questions generated from partial context can prepare learners for the full exam.


\paragraph{Forming the Exam.}
We extract exam questions to form $E$ from each chapter’s end-of-chapter materials. 
To ensure sufficient exam coverage while maintaining computational feasibility, we retain only chapters with at least 10 exam questions and cap each set to a maximum of 25.
We also use an LM to map each exam question to the sections relevant for answering it in order to use this information for baselines such as few-shot and supervised fine-tuning (SFT). 
To analyze the cognitive demands of the exam questions, we annotate each \(e_j \in E\) with a Bloom’s taxonomy category~\cite{krathwohl2002revision_bloom} using an LM. 
Refer to Appendix~\ref{appdx:bloom-taxonomy-level-distribution} for details on question type diversity in \ourdata according to Bloom's taxonomy.


\paragraph{Train-Test Split.}
For each subject, we select 25 chapters $C$ arranged in natural curriculum order: the first 20 for training and the last 5 for evaluation. 
Preserving sequential structure helps avoid leakage of prerequisite knowledge and supports realistic evaluation of learning progression.

\begin{table}[t]
\small
    \centering
    \resizebox{\columnwidth}{!}{
        \begin{tabular}{lccccc}
        \toprule
            \textbf{Subject} & \textbf{\# $C$} & \textbf{Split} & \textbf{\# $E$ / $C$} & \textbf{\% $E$ w/ A} & \textbf{\# $S$ / $C$} \\
        \midrule
            Microbiology & 20 & Train & 12.4 & 64\% & 16.4 \\
                         & 5  & Test  & 13.4 & 58\% & 17.0 \\
        \midrule
            Chemistry    & 20 & Train & 14.2 & 51\% & 11.0 \\
                         & 5  & Test  & 16.2 & 49\% & 6.4 \\
        \midrule
            Economics    & 20 & Train & 12.2 & 23\% & 14.1 \\
                         & 5  & Test  & 12.2 & 23\% & 14.4 \\
        \midrule
            Sociology    & 20 & Train & 10.4 & 62\% & 16.6 \\
                         & 5  & Test  & 11.2 & 67\% & 19.0 \\
        \midrule
            US History   & 20 & Train & 7.2 & 51\% & 14.9 \\
                         & 5  & Test  & 8.4 & 38\% & 13.2 \\
        \bottomrule
        \end{tabular}
    }
    \caption{\textbf{Data Statistics} of \ourdata. \# $ C$: number of chapters, \# $E/C$: avg. number of questions per chapter, \% $E$ w/ A: proportion of questions that have a reference answer, \# $S/C$: avg. number of sections per chapter.}
    \label{tab:textbook-exam-statistics}
    \vspace{-0.5cm}
\end{table}

\subsection{Data Statistics}
\label{ssec:textbook-exam-stat}

The statistics for \ourdata resulting from data processing are shown in Table \ref{tab:textbook-exam-statistics}. The general distributions of the numbers of questions and sections per chapter are similar between the training and test sets. 
Refer to Appendix~\ref{appendix:data_processing} for further details on \ourdata.

\section{Experimental Setup}
\label{sec:experiment-setup}

To evaluate \ours, we compare a variety of question generation (QG) methods and analyze how well \textit{utility} correlates with other question quality metrics (\textit{i.e.,} \textit{salience}, \textit{expected information gain}).


\subsection{Baselines}

\paragraph{No-Study baseline.}
$s_\emptyset$ denotes the exam score of a simulated learner without any study material, reflecting prior knowledge and serving as a lower bound.

\paragraph{Prompting baselines.}
\begin{itemize}
    \item \textbf{Zero-shot}: Prompts the base model to generate a question that helps a student understand a given document section \(S_k\), without providing any examples.
    
    \item \textbf{Few-shot}: Augments the zero-shot prompt with five (section, exam question) pairs randomly sampled from the training set.
    
    \item \textbf{Chain-of-Thought (CoT)}: Similar to zero-shot, but the model is asked to first generate a reasoning trace before writing the question~\cite{wei2022chain, zhou2024self}.
    
    \item \textbf{Bloom-based}: Samples a Bloom's Taxonomy level~\cite{krathwohl2002revision_bloom} from the distribution observed in the training set (\autoref{fig:bloom-distribution}), then prompts the model to generate a question at that cognitive level for the current section.
\end{itemize}


\paragraph{Fine-tuning baselines.}
\begin{itemize}
    \item \textbf{Supervised Fine-tuning (SFT)}: fine-tunes the base model to predict exam questions given (section, exam question) pairs.
    \item \textbf{\ours (Ours)}: applies rejection sampling to fine-tune the model on only high-utility QA pairs from simulations with the training data.
    We fix the utility threshold $\theta$ to 0.1 based on empirical performance observed on a development subject (i.e., Chemistry), and use this value consistently across all subjects.
\end{itemize}

For SFT and \ours, we experiment with two variants:  
(a) \textit{Subject-specific}: model is trained and evaluated on the same subject;  
(b) \textit{Cross-subject}: model is trained on all subjects and evaluated on each individually.

\subsection{Question Quality Metric}
To evaluate whether our measure of utility is a good indicator of question quality, we compare it to the  following alternative quality metrics:

\paragraph{Salience.} Measures a question's relevance to a document \( D \) by a LM~\cite{wu2024questions}.
It is rated on a Likert scale from 1 to 5: 1 indicates that the question for section \( k \) is unrelated to \( S_{[1:k]} \) and contributes minimally to understanding, while 5 indicates that it is essential to comprehending \( S_k \).
\paragraph{Expected Information Gain (EIG).} Quantifies the reduction in uncertainty about a student's knowledge state after answering a question~\cite{lindley1956measure, schaeffer2003science, rao-daume-iii-2018-learning, yu-etal-2020-interactive, white-etal-2021-open, keh-etal-2024-asking}.
We estimate EIG using an LM by computing the entropy of the model's token probability distribution before and after conditioning on the first token of the correct answer, capturing the shift in the model's belief distribution.


\subsection{Experiments and Implementation Details}  
All LMs used in data preprocessing
and throughout the framework, 
including the question generator \( M_q \), answer generator \( M_a \), learner \( M_l \), and evaluator \( M_e \) in the reader simulator, are based on \texttt{gpt-4o-mini}.
For SFT in the baseline and rejection sampling in \ours, we use OpenAI fine-tuning API to further train \texttt{gpt-4o-mini}.
We limit question generation to a single question per section and run three trials on the test set due to computational cost constraints of running inference and fine-tuning. Refer to Appendix \ref{appendix:cost-analysis} for a cost analysis of \ours.
For all other implementation details not covered here, refer to Appendix \ref{appdx:baseline-details} for details on baselines and \ref{appendix:evaluation-metrics} for details on question quality metrics.

\section{Experimental Results}

We empirically evaluate \ours along four dimensions: 
(1) overall exam performance; 
(2) relationship of utility with existing quality metrics and qualitative characteristics of high-utility questions; (3) the effects of design choices such as rejection threshold and model variants; and (4) human evaluation to validate \ours results.


\subsection{Overall Performance}
\label{ssec:overall-performance}
\begin{table*}[!t]
    \centering
    \resizebox{\textwidth}{!}{%
        \begin{tabular}{lccccc|c}
            \toprule
            & \textbf{Microbiology} & \textbf{Chemistry} & \textbf{Economics} & \textbf{Sociology} & \textbf{US History} & \textbf{Average} \\
            \midrule
            No-study $s_\emptyset$ & 45.8$_{1.0}$ & 9.2$_{0.9}$ & 0.3$_{1.1}$ & 61.2$_{0.8}$ & 3.2$_{1.2}$ & 23.9$_{0.7}$ \\
            \midrule
            Zero-shot & 62.3$_{0.4}$ (+16.5) & 39.7$_{0.3}$ (+30.5) & 40.3$_{0.5}$ (+40.0) & 60.7$_{0.2}$ (-0.5) & 19.4$_{0.1}$ (+16.2) & 44.5$_{0.4}$ (+20.6) \\
            Few-shot & 61.6$_{0.5}$ (+15.8) & \underline{45.6}$_{0.2}$ (+36.4) & \underline{46.8}$_{0.3}$ (+46.5) & 62.4$_{0.1}$ (+1.2) & 15.8$_{0.5}$ (+12.6) & 46.4$_{0.2}$ (+22.5) \\
            Chain-of-thought & 61.4$_{0.3}$ (+15.6) & 45.4$_{0.5}$ (+36.2) & 45.9$_{0.2}$ (+45.6) & 61.1$_{0.1}$ (-0.1) & 18.6$_{0.4}$ (+15.4) & 46.5$_{0.3}$ (+22.6) \\
            Bloom-based & 57.3$_{0.1}$ (+11.5) & 36.6$_{0.5}$ (+27.4) & 29.4$_{0.2}$ (+29.1) & 61.5$_{0.4}$ (+0.3) & 21.7$_{0.3}$ (+18.5) & 41.3$_{0.5}$ (+17.4) \\
            \midrule
            SFT (Subject-Specific) & 65.4$_{0.2}$ (+19.6) & 23.9$_{0.3}$ (+14.7) & 46.1$_{0.1}$ (+45.8) & \underline{64.2}$_{0.4}$ (+3.0) & 20.2$_{0.5}$ (+17.0) & 44.0$_{0.2}$ (+20.1) \\
            SFT (Cross-Subject) & 58.7$_{0.4}$ (+12.9) & 21.1$_{0.1}$ (+11.9) & \underline{46.8}$_{0.5}$ (+46.5) & 63.4$_{0.3}$ (+2.2) & \underline{26.3}$_{0.2}$ (+23.1) & 43.3$_{0.1}$ (+19.4) \\
            \midrule
            \textsc{QUEST} (Subject-Specific) & \bf 75.6$_{0.5}$$^*$ (+29.8) & \bf 45.7$_{0.4}$ (+36.5) & \bf 58.2$_{0.2}$$^*$ (+57.9) & \bf 64.9$_{0.3}$ (+3.7) & \bf 31.1$_{0.1}$$^*$ (+27.9) & \bf 55.1$_{0.5}$$^*$ (+31.2) \\
            \textsc{QUEST} (Cross-Subject) & \underline{73.3}$_{0.2}$$^*$ (+27.5) & 41.3$_{0.3}$ (+32.1) & \underline{46.8}$_{0.4}$ (+46.5) & \underline{64.8}$_{0.5}$ (+3.6) & 25.2$_{0.2}$ (+22.0) & \underline{50.3}$_{0.1}$$^*$ (+26.4) \\
            \bottomrule
        \end{tabular}
    }
    \caption{\textbf{End-of-chapter exam scores} for different question generation methods, averaged over three trials. 
    \ours achieves the highest scores across all subjects. 
    Gains (in parentheses) are relative to the no-study base score $s_\emptyset$. Standard errors are reported as subscripts.
    Top and second-best scores are \textbf{bolded} and \underline{underlined}. 
     $^*$Improvements over non-\ours baselines that are statistically significant at $p < 0.01$ with paired t-test.}
    \label{tab:question-gen-results}
\end{table*}

Table~\ref{tab:question-gen-results} reports the simulated \textit{exam scores} for different question generators.
Our main findings are:
(1) \textbf{Prompting methods} (Few-shot, CoT, Bloom-based) improve over the no-study baseline, leveraging better task understanding and reasoning~\cite{brown2020language, wei2022chain, zhou2024self}.
However, they do not explicitly optimize for learning outcomes, limiting their effectiveness;
(2) \textbf{SFT} shows mixed results—improving in some subjects but underperforming in others—and does not consistently outperform prompting on average; and
(3) \textbf{\ours} achieves the highest exam scores across all subjects, with an average gain of 18.5\% over the next strongest baseline (CoT).
Its subject-specific variant consistently outperforms the cross-subject version, highlighting the importance of domain-specific optimization.
These results show that outcome-based filtering and training yields the most effective instructional questions.


\subsection{Question Utility Analysis}
\label{ssec:evaluation-metrics}
\paragraph{Correlation with indirect metrics.}
To analyze whether indirect metrics align with our proposed \textit{utility}, we compute all three metrics (\textit{utility}, \textit{salience}, and \textit{expected information gain (EIG)}) on generated questions from the training set.
As shown in Table~\ref{tab:correlation_results}, utility has only a weak correlation with salience ($\rho = 0.097$) and no significant correlation with EIG ($\rho = -0.022$). 

\begin{table}[t]
    \centering
    \resizebox{\columnwidth}{!}{%
        \begin{tabular}{ll
        rr}
            \toprule
            Metric 1 & Metric 2 & \textbf{Correlation $\rho$} & \textbf{$p$-value} \\
            \midrule
            \multirow{5}{*}{\textbf{Utility}} & Salience & 0.097 & 0.003 \\
            & EIG  & $-0.022$ & 0.512 \\
            & Exam Embedding Sim. & 0.250 & 0.001 \\
            & Exam ROUGE & 0.040 & 0.010 \\
            & Bloom's Taxonomy & 0.120 & 0.001 \\
            \bottomrule
        \end{tabular}
    }
    \caption{\textbf{Correlation} between utility and other metrics. 
    }
    \vspace{-0.2cm}
    \label{tab:correlation_results}
\end{table} 
This suggests that utility captures a distinct signal from existing indirect metrics, which may not reliably reflect a question’s actual impact on learning outcomes.
Table~\ref{tab:utility_examples_excerpt} makes that gap concrete for Chemistry: a high-salience, high-utility question asks how combining elements yields new properties; a low-salience, high-utility question asks for a specific list of green-chemistry principles that happens to help on the exam; and a high-salience, low-utility question asks a generic definition of ionic charge.
The same High/High, Low/High, and High/Low pattern appears across subjects (Appendix~\ref{appendix:qualitative-examples}), with no consistent surface-level cue that separates high- from low-utility questions.

\begin{table*}[t]
    \centering
    \small
    \begin{tabularx}{\textwidth}{@{}llX@{}}
        \toprule
        \textbf{Sal.} & \textbf{Util.} & \textbf{Question} \\
        \midrule
        High & High & How does the combination of elements result in properties that are different from those of the individual elements in their uncombined states, and can you provide more examples of such transformations? \\[0.4em]
        Low & High & What are the `Twelve Principles of Green Chemistry'? \\[0.4em]
        High & Low & What happens to the charge of an atom when it gains or loses electrons? \\
        \bottomrule
    \end{tabularx}
    \caption{\textbf{Chemistry examples} of generated questions by salience and utility. Additional subjects are in Appendix~\ref{appendix:qualitative-examples}.}
    \label{tab:utility_examples_excerpt}
\end{table*}


\begin{table*}[!t]
    \centering
    \small
    \resizebox{\textwidth}{!}{%
        \begin{tabular}{lrrr rrr rrr rrr rrr}
            \toprule
            & \multicolumn{3}{c}{\textbf{Microbiology}} & \multicolumn{3}{c}{\textbf{Chemistry}} & \multicolumn{3}{c}{\textbf{Economics}} & \multicolumn{3}{c}{\textbf{Sociology}} & \multicolumn{3}{c}{\textbf{US History}} \\
            \cmidrule(lr){2-4} \cmidrule(lr){5-7} \cmidrule(lr){8-10} \cmidrule(lr){11-13} \cmidrule(lr){14-16}
            \textbf{Filter} & \textbf{E} & \textbf{Sal.} & \textbf{EIG} & \textbf{E} & \textbf{Sal.} & \textbf{EIG} & \textbf{E} & \textbf{Sal.} & \textbf{EIG} & \textbf{E} & \textbf{Sal.} & \textbf{EIG} & \textbf{E} & \textbf{Sal.} & \textbf{EIG} \\
            \midrule
            $utility > 0.1$ & \textbf{76} & 4.27 & $-$.18 & \textbf{46} & \textbf{4.65} & $-$.20 & \textbf{58} & \textbf{4.70} & $-$.04 & \textbf{65} & \textbf{4.49} & $-$.02 & \textbf{31} & 4.65 & \bf $-$.01 \\
            $salience = 5$  & 73 & \textbf{4.42} & $-$.24 & 39 & 4.46 & $-$.22 & 46 & 4.66 & $-$.08 & 64 & \textbf{4.49} & $-$.03 & 23 & \textbf{4.68} & $-$.02 \\
            $EIG > 0$      & 61 & 4.21 & \textbf{$-$.17} & 32 & 4.40 & \textbf{$-$.09} & 47 & 4.65 & \textbf{.01} & 62 & 4.46 & \textbf{.01} & 21 & 4.65 & \bf $-$.01 \\
            \bottomrule
        \end{tabular}
    }
    \caption{\textbf{Performance of \ours models trained on different filtering criteria.} 
    Each group reports: exam score (\textbf{E}), average salience (\textbf{Sal.}), and expected information gain (\textbf{EIG}).}
    \label{tab:quest-indirect}
\end{table*}

\paragraph{Overlap with exam questions.}
To evaluate the relationship between generated high-utility questions and exam questions, we measure their semantic and lexical similarity.
For each generated question, we compute cosine similarity using embeddings with \texttt{text-embedding-3-small} from OpenAI for semantic overlap and the ROUGE score~\cite{lin2004rouge} for lexical overlap with all exam questions in the same chapter.
We then assess the correlation between utility and the most similar exam question based on these measures.
The correlation between utility and semantic similarity is 0.25 ($p < 0.001$), indicating a weak positive relationship, while the correlation with ROUGE is nearly zero at 0.04 ($p < 0.01$).
These findings suggest that high-utility questions are not simple paraphrases of exam questions but may introduce novel concepts that enhance learning beyond surface-level similarity.

\paragraph{Bloom's taxonomy analysis.}
An interesting observation is that Bloom's taxonomy, which categorizes cognitive depth based on question type—where ``what'' questions typically involve simple recall, while ``why'' and ``how'' questions require deeper processing—also does not strongly correlate with utility.
Using Bloom's taxonomy as a cognitive depth scale (Likert 1-6), the correlation between utility and cognitive depth is 0.12 ($p < 0.001$), indicating a weak positive relationship.

\paragraph{Optimization on indirect metrics.}
To assess whether indirect metrics can serve as effective training objectives, we compare \ours models trained on datasets filtered by three different criteria:
(1) questions with \textit{utility} $> 0.1$ (Ours),
(2) questions with \textit{salience} = 5, and
(3) questions with \textit{EIG} $> 0$.
We evaluate each model's output using three metrics: \textit{exam score} (utility), average salience, and average EIG. 
Results are shown in Table~\ref{tab:quest-indirect}.

We observe that training on utility-optimized questions consistently yields the highest exam performance across all subjects.
In contrast, optimizing for salience or EIG improves the respective metric, but fails to enhance utility.

These results underscore a key distinction: indirect metrics may capture intrinsic quality (\textit{e.g.,} relevance or informativeness), but do not translate into improved downstream outcomes. 
In contrast, direct optimization on utility, measured via simulated exam performance by modeling a learner's mental model, leads to more effective question generation for downstream outcomes.

\begin{figure}[!t]
    \centering
    \begin{minipage}{\columnwidth}
    \centering
    \includegraphics[width=\columnwidth]{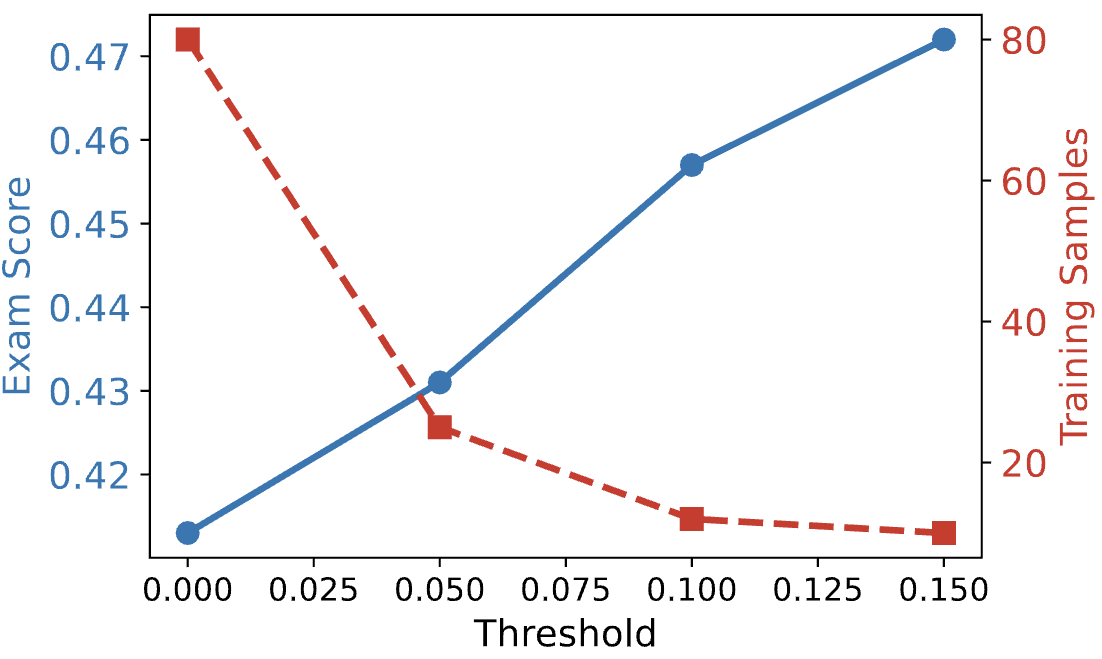}
    \end{minipage}
    \caption{\textbf{Impact of threshold} in \ours on exam scores for Chemistry.
    }
    \label{fig:rs_analysis}
    \vspace{-0.5cm}
\end{figure}

\subsection{Rejection Sampling Analysis}
\label{ssec:rs-analysis}

We examine the effect of the utility threshold on rejection sampling results by varying it for chemistry as a representative example. 
Figure~\ref{fig:rs_analysis} illustrates that increasing the utility threshold leads to generated questions that enable higher exam scores, confirming that stricter filtering selects more educationally valuable questions.
However, higher thresholds also reduce the number of training examples, potentially limiting model learning. 
This trade-off highlights the importance of balancing quality and quantity.
The results suggest that expanding the training pool by sampling more questions per section while maintaining a high threshold may yield further improvements.

\subsection{Model Variants}
\label{ssec:model-variants}

\begin{table*}[t]
    \centering
    \small
    \resizebox{\textwidth}{!}{%
        \begin{tabular}{lcccccccc}
            \toprule
            & \textbf{QG ($M_q$)} & \textbf{AG ($M_a$)} & \textbf{RS ($M_l$)} & \textbf{Microbiology} & \textbf{Chemistry} & \textbf{Economics} & \textbf{Sociology} & \textbf{US History} \\
            \midrule
            \multirow{4}{*}{Zero-shot}
            & \texttt{4o-mini} & \texttt{4o-mini} & \texttt{4o-mini} & 62.0 & 41.4 & 39.8 & 60.9 & 23.3 \\
            & \texttt{4o} & \texttt{4o-mini} & \texttt{4o-mini} & 68.1 & 45.7 & 46.6 & 63.4 & 18.0 \\
            & \texttt{4o-mini} & \texttt{4o} & \texttt{4o-mini} & 68.2 & 42.2 & 48.0 & 63.4 & 23.2 \\
            & \texttt{4o-mini} & \texttt{4o-mini} & \texttt{4o} & 71.0 & 17.3 & 47.6 & 56.4 & 26.3 \\
            \midrule
            \ours & \texttt{4o-mini} & \texttt{4o-mini} & \texttt{4o-mini} & \bf 75.6 & \bf 45.7 & \bf 58.2 & \bf 64.9 & \bf 31.1 \\
            \bottomrule
        \end{tabular}
    }
    \caption{\textbf{Model-size ablations} of \ours modules on \ourdata. Each zero-shot row changes one of $M_q$, $M_a$, or $M_l$ while holding the others fixed. \ours remains strongest across subjects.}
    \label{tab:model-variants}
\end{table*}

To test whether \ours's gains depend on a particular model size, we vary the model used for the question generator ($M_q$), answer generator ($M_a$), and reader simulator ($M_l$), changing one module at a time in the zero-shot setting.
For the $M_a$ and $M_l$ variants, we reuse the same questions (and answers) from the zero-shot baseline and re-run only the affected stage.
Results are shown in Table~\ref{tab:model-variants}.

Upgrading $M_q$ or $M_a$ to \texttt{gpt-4o} yields moderate gains (5.7\% and 7.1\% on average), but both still trail the \ours-trained \texttt{gpt-4o-mini} generator by more than 11\%.
Using \texttt{gpt-4o} as the reader simulator is mixed: some subjects improve, while Chemistry drops sharply, suggesting that a larger learner can change scoring behavior rather than uniformly raise exam scores.
Overall, substituting a stronger model into a single module does not replace training on high-utility questions.

\subsection{Human Evaluation}
\label{ssec:human-evaluation}

The validity of \ours results, and thus \ours's practical value of estimating question utility and using it to improve question generation, hinges on whether utility reflects actual usefulness for downstream outcomes perceived by humans, and that it is distinct from other metrics such as salience and EIG. 
To validate that utility is indeed a reliable approximation of such usefulness, we conduct a human evaluation study
\footnote{The human evaluation study is available at \url{https://quest-human-eval.streamlit.app/}.}
where human participants take on the role of a learner in \ours.
For a single evaluation sample, we ask human participants to answer a set of related exam problems after reading a section of a chapter and a set of questions, one of each with exclusively high utility, high salience, and high EIG, and their corresponding answers.
Next, they are instructed to rank questions without ties based on their usefulness for solving the exam problems and also rate their usefulness and interestingness (i.e. how intriguing the question is regardless of its usefulness) on a Likert scale from 1 to 5, and provide a brief explanation for their rankings and ratings.

This study is construct validation of the utility metric, not a head-to-head comparison of generators.
An ideal check would compare \ours-trained models against every baseline with human learners across many sections and subjects, but that design is impractical: solving exam items after multiple QA sets is cognitively demanding, so session length, fatigue, and variance grow quickly, and fair comparisons require counterbalancing that multiplies conditions.
Human sessions also cannot score a full exam the way \ours{} does; they must be limited to a few items.
We therefore test whether simulation-computed utility predicts human-perceived usefulness across models and subjects.
The results are shown in Table~\ref{tab:human-evaluation}.

\begin{table}[t]
    \centering
    \resizebox{\columnwidth}{!}{%
        \begin{tabular}{llll}
            \toprule
            & EIG & Salience & Utility \\ 
            \midrule 
            $\rho$ Useful. $\uparrow$ & $-$0.219 & $-$0.007 & \bf 0.321 \\ 
            \midrule 
            $\mu$ Useful. $\uparrow$ & 3.00$_{0.10}$ & 2.97$_{0.09}$ & \bf 3.96$_{0.09}$$^{**}$ \\ 
            $\mu$ Interesting. $\uparrow$ & 3.13$_{0.08}$ & 3.23$_{0.09}$ & 3.26$_{0.09}$ \\ 
            $\mu$ Rank $\downarrow$ & 2.25$_{0.06}$ & 2.04$_{0.05}$ & \bf 1.71$_{0.05}$$^{*}$ \\ 
            \bottomrule
        \end{tabular}
    }

    \caption{\textbf{Human evaluation results.} $\mu$ denotes the mean and $\rho$ denotes correlation. Standard errors are reported as subscripts. $^{*}$ and $^{**}$ denote statistically significant at $p < 0.05$ and $p < 0.01$ with paired t-test, respectively.}
    \vspace{-0.2cm}
    \label{tab:human-evaluation}
\end{table}

The results show that utility is moderately positively correlated with usefulness perceived by humans ($\rho = 0.321$), while other metrics are weakly negatively correlated.
In addition, high utility questions are consistently ranked and rated higher than high salience and high EIG questions for usefulness. 
These results establish that utility as measured by LM simulations can capture usefulness perceived by humans to some extent.
It is also worth noting that high utility questions are rated as interesting as high salience and high EIG questions, indicating that optimizing for utility does not necessarily lead to a tradeoff with interestingness.
Participant instructions, exclusivity sampling, and the evaluation interface are in Appendix~\ref{appendix:human-evaluation-details}. 





\section{Related Work}

\paragraph{Question Generation.}
Recent studies have explored generating \textit{information-seeking} questions to enhance comprehension by reflecting the inquisitive nature of human question-asking.
These include curiosity-driven inquisitive questions~\cite{ko-etal-2020-inquisitive, wu2024questions}, confusion-driven inquisitive questions~\cite{chang2024booookscore}, follow-up questions~\cite{meng-etal-2023-followupqg}, and clarifying questions~\cite{chen2018learningq, rao-daume-iii-2018-learning, rao-daume-iii-2019-answer, kumar-black-2020-clarq, majumder-etal-2021-ask}.
A widely adopted framework in this area is the concept of Questions Under Discussion (QUD), which views each sentence as the answer to an implicit or explicit question from prior context~\cite{van1995discourse, roberts2012information, onea2016potential, benz2017questions}.
Building on this framework, recent works have applied question generation to tasks such as decontextualization~\cite{newman-etal-2023-question}, which recovers missing context to make snippets standalone, and elaboration~\cite{wu-etal-2023-elaborative}, which generates additional details to enhance clarity. 
Other applications include discourse comprehension by linking sentences to their broader context~\cite{ko-etal-2022-discourse}, 
planning for summarization~\cite{narayan2023conditional}, 
and modeling information loss during text simplification~\cite{trienes-etal-2024-infolossqa}.

\paragraph{Question Evaluation.}
Existing work evaluates information-seeking questions based on expected information gain, measuring the additional information provided by the answers~\cite{lindley1956measure, schaeffer2003science, rao-daume-iii-2018-learning, yu-etal-2020-interactive, white-etal-2021-open, keh-etal-2024-asking}, or salience, assessing a question's relevance and importance in introducing new concepts or clarifying key ideas~\cite{wu2024questions}.
Empirical findings show that QUDs—questions guaranteed to be addressed in the article—consistently receive high salience ratings~\cite{ko-etal-2022-discourse}, reflecting reader expectations and linking to the utility of the question~\cite{van2003questioning, wu2024questions}.
In contrast, our work directly evaluates the utility of questions using an LM as a human simulator.

Another prominent line of work that evaluates question quality is test information functions in item response theory~\cite{harvey1999item, sumita-etal-2005-measuring, lalor-etal-2016-building, zhou-etal-2019-intelligent}, but its primary concern is measuring how well existing exam questions can differentiate learners of varying proficiency or estimating latent traits across learners, not the \textit{interventional} effect of a question asked while learning on improving downstream performance.

\paragraph{LM as Human Simulation.}
LMs have been used to simulate real-world social interactions, starting with everyday human interactions~\cite{park2024generative, maharana-etal-2024-evaluating, lee2025realtalk} and expanding to controlled environments such as market competition~\cite{zhaocompeteai}, hospital settings where medical staff and patients engage in treatment simulations~\cite{li2024agent, schmidgall2024agentclinic}, news interviews~\cite{lu2024newsinterviewdatasetplaygroundevaluate}, content moderation~\cite{cho-etal-2024-language, shaikh2024rehearsal}, and academic peer review~\cite{jin2024agentreview}.
In education, LMs simulate classroom interactions between teachers and students~\cite{zhang2024simulating} and predict learning outcomes to improve educational materials~\cite{he2024evaluating, zelikman-etal-2023-generating}.
Our work builds on \citet{he2024evaluating}, using simulated learning outcomes as an evaluation metric for question quality and as a reward signal to improve question generation.

\section{Conclusion}

We introduce \ours, a framework for estimating question utility by measuring the question's impact on a simulated learner's performance on real exams and fine-tuning models to generate high-utility questions. 
We establish that question utility reflects usefulness perceived by humans to some extent, establishing \ours's practical value of estimating question utility and using it to improve question generation.
We show that \ours-trained models consistently outperform existing prompting and fine-tuning baselines.
On the other hand, indirect metrics from previous work such as salience and expected information gain fail to capture actual learning outcomes, emphasizing the need for evaluation grounded in task-specific goals.


\section*{Limitations}

We acknowledge that our scope of question utility and \ours studied in this work is confined to an educational setting with textbooks that contain learning material with corresponding exam questions. 
However, the definition of question utility as the contribution a question and its resulting answer makes to improved outcomes for a downstream task is generalizable and can be applied to other scenarios where downstream performance can be concretely quantified.

One of the main technical limitations of \ours is that we generate only one question per section due to budget constraints. 
Due to the stochastic nature of language models, the type of questions that get generated for each section may vary significantly. 
However, we believe this effect to be minimal and the variance to be sufficiently captured in our results, as the improvements of \ours over other baselines were statistically significant at $p<0.01$ on average and for three of the five subjects in \ourdata, as reported in \autoref{tab:question-gen-results}. 
Due to the same constraints, we have not tested \ours with rejection sampling that generates more than one question per section. 
However, the threshold analysis results shown in \autoref{fig:rs_analysis} indicate that setting a higher threshold leads to larger improvements, and therefore sampling a larger set of questions from each section will most likely lead to further improvements for \ours. 
Since \ours is effective even with the minimal setup of generating a single question per section, the main value of our work remains valid in that we have conceptually demonstrated that rewards from LM-based simulations can lead to meaningful improvements over baselines. 

A major aspect overlooked by our work is the effect of a learner's persona and prior knowledge on the utility of a question. 
There is a wide variety in learning styles and the extent of prior knowledge that a learner may have, which all influence the utility of a question. 
As a preliminary exploration, we have kept the learner persona and prior knowledge fixed for all experiments. 
We leave it to future work to explore the effect of varying learner personas and prior knowledge on measuring question utility. 

Lastly, \ours simulations incur a significant number of API calls, which can be costly and time-consuming, making it difficult to scale when hyperparameters such as the number of questions and exam problems become large. 
However, as shown in our cost analysis in Appendix \ref{appendix:cost-analysis}, the cost is still relatively affordable given the high number of API calls and total tokens when using \texttt{gpt-4o-mini}. 
In addition, with the rapid improvements in language model inference efficiency, costs and time for \ours simulations will continue to decrease.
Most importantly, we argue that \ours enables a scalable and controlled experimental setup to measure outcome-driven metrics that is otherwise impractical in the real world and have shown potential for practical value via human evaluation that the utility of a question as measured by \ours reflects actual usefulness as perceived by humans to some extent.
With further improvements in language model capabilities for simulating human behavior, frameworks like \ours will become more reliable and applicable to more complicated and nuanced use cases.
We hope this work inspires more work that grounds utility in real human outcomes for complex tasks that were previously infeasible to model.

\section*{Ethical Considerations}

Although our work has shown that LM-based simulators can provide an effective reward signal for training better question generators for a specific downstream task and that this reward signal reflects actual usefulness as perceived by humans, we make no claims to how \ours correctly simulates a real human being's learning process and behavior. 
Human learning is a complex process involving long-term memory, context, and numerous other factors that are not captured by \ours.
Therefore, we do not advocate using our measure of question utility beyond a proof of concept for developing metrics grounded in downstream outcomes. 
In other words, \ours should not be used to directly assess the quality of questions that students ask in a classroom setting and provide feedback based on its assessment. 

The main reason is that the core assumption of our work -- that the exam problems contained in \ourdata suitably evaluate a learner's understanding of the learning material -- is not guaranteed to hold in other settings.
The utility of a question as defined by \ours is heavily dependent on what the exam problems are, and if the exam questions are misaligned with desirable educational outcomes, e.g., exam questions that only require rote memorization rather than critical thinking, a student's question may be considered low utility despite being useful in other scenarios, such as brainstorming.
While we take extensive measures to ensure that this is not the case in our experiments given the wide variety of questions across subjects and types of questions included in \ourdata as shown in \autoref{fig:bloom-distribution}, we cannot guarantee similar diversity and comprehensiveness in other textbook datasets. 

\section*{Acknowledgments}
This work was funded by the Defense Advanced Research Projects Agency with award HR00112220046.
We thank the volunteer participants who contributed to our human evaluation.

\bibliography{anthology,custom}

\appendix
\widowpenalty=10000
\clubpenalty=10000
\displaywidowpenalty=10000

\section{\ours Details}
\label{appendix:quest-details}
\ours consists of four key modules. 
(1) The \textbf{Question Generator} (\(M_q\)) generates a set of questions \( Q = \{q_1, q_2, \dots, q_n\} \) from a document \( D \) (Section 2.3);
(2) The \textbf{Answer Generator} (\(M_a\)) then produces corresponding answers, forming question-answer pairs \( \text{QA} = \{(q_1, a_1), (q_2, a_2), \dots, (q_n, a_n)\} \) using parametric knowledge (Section 2.3).
(3) The \textbf{Reader Simulator} consists of a \textbf{Learner} (\(M_l\)) and an \textbf{Evaluator} (\(M_e\)). The learner model \(M_l\) simulates a learner’s understanding by attempting the final exam using only the generated QA pairs, formulated as \( P \sim M_l(E; \text{QA}) \). 
The evaluator model \(M_e\) then assesses the learner’s responses \( P \) by comparing them against ground-truth answers when available or using parametric knowledge to assign a score.
In this section, we present the prompts that we use for each of the components in \ours and validate their reliability for the role that they simulate. 

 
\subsection{Question Generator}
\label{appdx:question-generator}

\begin{tcolorbox}[
title=Question Generator, myboxstyle]
\texttt{Article:} \(\mathit{S_{[1:k-1]}}\)\\
\texttt{Student is currently reading the section:} \(\mathit{S_k}\).
\begin{lstlisting}
Generate a question that helps the student
understand the section better.

Output in the following JSON format:
```json
{"question": question}
```
\end{lstlisting}
\end{tcolorbox}

\subsection{Answer Generator}
\begin{tcolorbox}[title=Answer Generator, myboxstyle]
\begin{lstlisting}
questions: {{questions}}

Answer each question shortly and output
in following JSON format:
```json
{"qa_pairs": [
    {"question": question_1, "answer": answer_1},
    {"question": question_2, "answer": answer_2},
    ...
    {"question": question_n, "answer": answer_n}
]}
```
\end{lstlisting}
\end{tcolorbox}

We manually evaluate whether the answers generated by our answer generator are mostly correct so that they can be used as reliable information for taking the exam $E$. 
We inspect 20 random samples of generated questions for each subject for 100 total samples, and find that the answer generator provides factually correct answers to all of them, albeit with some cases not being completely exhaustive. 
For example, for the question \textit{``What are some examples of chemical properties, and how do they differ from physical properties in terms of changes in matter?''}, the answer is not an exhaustive list of all chemical properties, but the examples provided are all correct.
The general correctness is not surprising given that the textbooks are for entry-level college courses and most subjects do not involve subjectivity.

\subsection{Learner}
\begin{tcolorbox}[title=Learner, myboxstyle, breakable]
\begin{lstlisting}
You are now a learner participating in a structured learning simulation.
Your task is to:

1. **Study the Provided Learning Materials:** Carefully read and understand
   the content enclosed in the [LEARNING MATERIALS] tags.

2. **Answer the Exam Questions Using Only the Learning Materials:**
   When you respond to the questions in the [EXAM], you must:
   - Base all answers solely on the information contained in the
     [LEARNING MATERIALS].
   - Clearly show how your reasoning follows from the [LEARNING MATERIALS].
   - If the question asks about something not covered in the
     [LEARNING MATERIALS], do not provide an answer or guess. Instead,
     respond exactly with:

     I don't know. I have not studied this.

   - Do not use information from outside the [LEARNING MATERIALS].

3. **No External Knowledge or Guessing:**
   Provide no additional reasoning or information if the content is not in
   the [LEARNING MATERIALS].

Let's begin.

[LEARNING MATERIALS]
{{learning_material}}
[/LEARNING MATERIALS]

Now, proceed to the exam below and answer as instructed:

[EXAM]
{{exam}}
[/EXAM]

Respond with answers in the following JSON format (key: exam question number;
value: your answer):
{
    "1": "< your answer to exam question 1 >",
    "2": "< your answer to exam question 2 >",
    ...
}

# TARGET TEXTBOOK CONTENT:
{{ target_content }}
\end{lstlisting}
\end{tcolorbox}

This is the main component of \ours that is responsible for measuring the utility of a question and that we validate with human evaluators, results shown in Section \ref{ssec:human-evaluation} and details of the study in Appendix \ref{appendix:human-evaluation-details}.

\subsection{Evaluator}

\begin{tcolorbox}[title=Evaluator, myboxstyle, breakable]
\begin{lstlisting}
You are a teacher who is evaluating a student's understanding of a document.

Here is the document:
{{document}}

Now, determine the correctness of the student's answers to the following
question.

question:
{{question}}

ground truth:
{{answer}}

student's answer:
{{prediction}}

Please provide a score between 0 and 1, where:
- 0 indicates the student's answer is completely incorrect.
- 1 indicates the student's answer is completely correct.

If ground truth is not provided (e.g., None), determine the correctness of
the student's answer based on your own understanding of the document.

Answer in the following JSON format:
{"score": <score>, "feedback": "<feedback>"}
\end{lstlisting}
\end{tcolorbox}

To validate whether the evaluator provides a reasonable assessment for the correctness of the answer, we manually examine 50 scores from each subject assigned to an answer when compared to the ground truth or based on the textbook content when one is not provided. 
Multiple choice or short answer questions are easy to assess, but answers to long free-form questions are more subjective and some questions involve multiple sub-questions.
Despite the variety of question types, we find that our prompt works well with \texttt{gpt-4o-mini} in assigning appropriate credit to each answer. 
Out of 250 scores we examine, there are only two instances for the sociology subject that incorrectly assigned full credit for a wrong answer: subculture was given full credit when the correct answer is counterculture for the question \textit{``The Ku Klux Klan is an example of what part of culture? 1. Counterculture 2. Subculture 3. Multiculturalism 4. Afrocentricity''}.
Answers that contain multiple items are also appropriately weighted in most cases. 
For example, for the question \textit{``What evidence would you use to support this statement: Ancient people thought that disease was transmitted by things they could not see.''}, the answer \textit{``Ancient texts from civilizations such as the Egyptians, Greeks, and Romans indicate an awareness of contagious diseases. For example, Hippocrates noted the spread of illnesses in crowded areas, and the Romans implemented public health measures during plagues.''} is given a score of 0.8 because the statement about Romans implementing quarantine is an overstatement; they took public health measures but not to an extent comparable to modern quarantine practices. 
These results motivate using LMs as automated answer generators and scorers for \ours.

\subsection{Human Evaluation Details}
\label{appendix:human-evaluation-details}

\begin{figure*}[t]
    \centering
    \includegraphics[width=\textwidth]{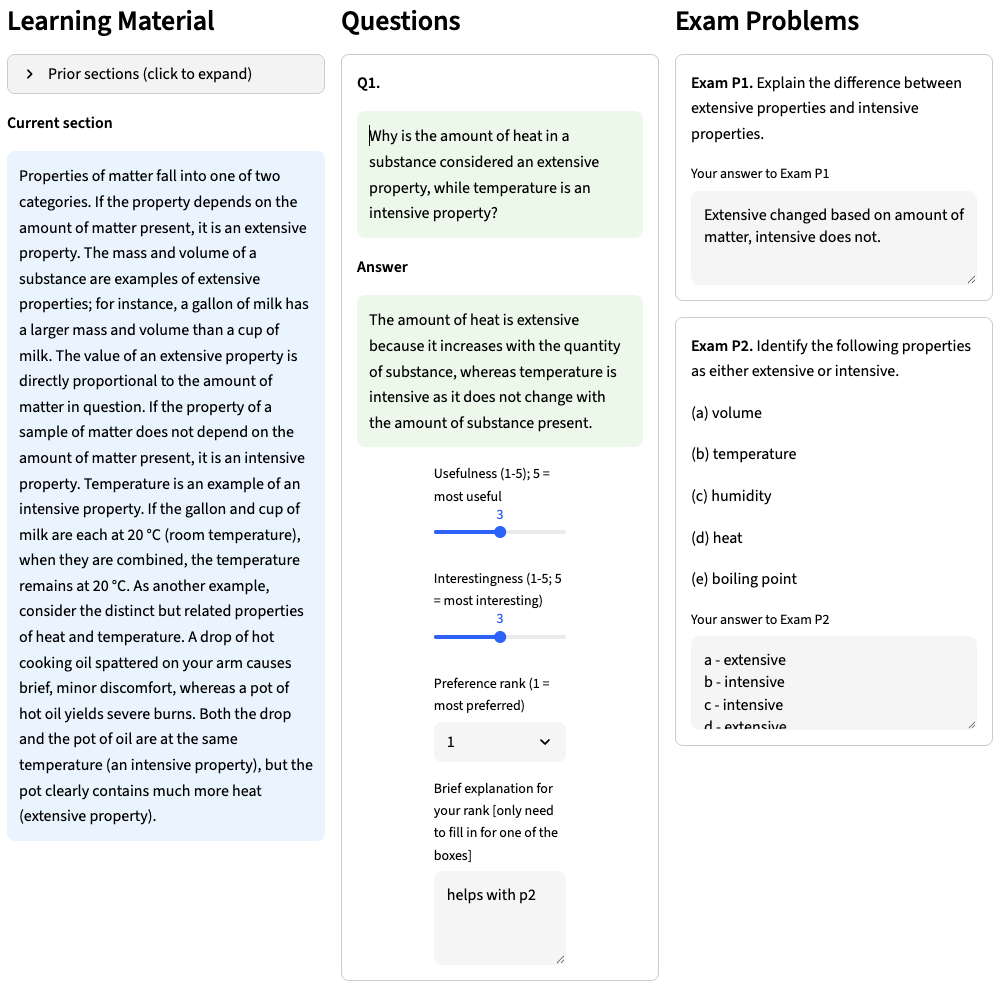}
    \caption{Human evaluation interface.}
    \label{fig:human-evaluation-interface}
\end{figure*}

\begin{figure*}[t]
    \centering
    \includegraphics[width=\textwidth,height=0.38\textheight,keepaspectratio]{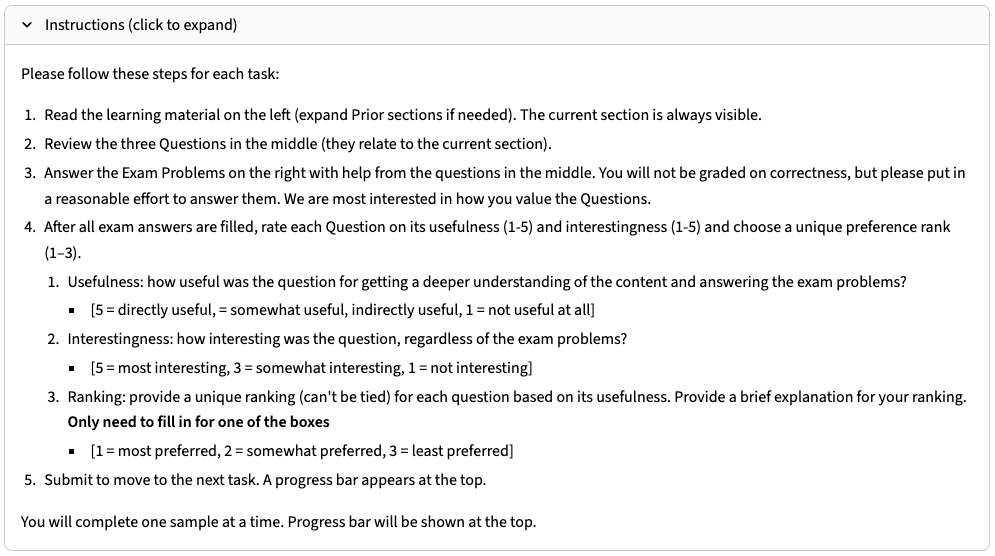}
    \caption{Human evaluation instructions.}
    \label{fig:human-evaluation-instructions}
\end{figure*}

\paragraph{Setup.}
The instructions for our human evaluation study are shown in \autoref{fig:human-evaluation-instructions} and the interface is shown in \autoref{fig:human-evaluation-interface}. 
Our human evaluation study is available at \url{https://quest-human-eval.streamlit.app/}.
We recruit eight volunteer participants from the University of Southern California, all located in North America, to each complete 19 samples that span the five subjects.
All participants provided informed consent before taking part.
Participants span a wide variety of demographics, but are all fluent in English and have a computer science background.
Participants are instructed to mark questions that they have prior knowledge of the answer to so that they are excluded from the evaluation.
Each sample consists of a section of a chapter, prior sections that can be opened to read if desired, and a set of questions and their corresponding answers, one of each with exclusively high utility, high salience, and high EIG.
Exclusivity is enforced by making sure that a high utility question is not also a high salience or high EIG question such that its salience and EIG scores are not within the top 20\% of the distribution. 
In practice, we find that questions with exclusively high utility, high salience, and high EIG are rare, mostly because salience is often high for most questions. 
Therefore, it requires nontrivial amount of sampling to get a set of questions that fulfill the exclusivity requirement, hence the relatively small number of evaluation samples. 
However, meaningfully replicating the learner simulation with human participants is a relatively cognitively loaded task that is not practical for a large number of samples.

With this setup, we collect a total of 107 samples, which we find to be sufficient for our analysis in getting statistically significant results for usefulness and ranking. 

The rationale for treating this study as construct validation of utility, rather than a head-to-head comparison of generators, is discussed in \secref{ssec:human-evaluation}.

\subsection{Cost Analysis}
\label{appendix:cost-analysis}

\ours requires generating questions that a simulated learner asks while reading through a textbook chapter and then approximating the learner's performance on related exam problems for various subsets of generated questions, which requires generating answers that models the learner's understanding and verifying their correctness. 
Therefore, the total number of input and output tokens for \ours can get costly, but we argue that it is not impractical when using a smaller model with lower cost and worthwhile given that \ours enables measuring a metric that is otherwise infeasible to measure. 
We provide the exact costs for using \texttt{gpt-4o-mini} for representative subjects in \autoref{tab:cost-analysis} (Microbiology as the most expensive and U.S. History as the least expensive).
Although the number of API calls is high and the number of total tokens is substantial, the cost is still relatively affordable. 
In addition, many of the calls can be processed as a pipeline of parallel batch jobs (e.g., generating questions and answers for all sections, running the learner simulator for all subsets of questions, etc.), which can reduce the cost significantly.
The bulk of the total costs with experimentation is actually incurred by fine-tuning the model with a custom dataset, which is a one-time expense but costs on average \$5 per fine-tuning job. 

\begin{table*}[t]
    \small
    \centering
    \begin{tabularx}{\textwidth}{@{}l>{\centering\arraybackslash}X>{\centering\arraybackslash}X>{\centering\arraybackslash}X>{\centering\arraybackslash}X>{\centering\arraybackslash}X@{}}
            \toprule
            \textbf{Subject} & \textbf{Input Tokens} & \textbf{Output Tokens} & \textbf{Total Tokens} & \textbf{API Calls} & \textbf{Cost} \\
            \midrule 
            \textbf{US History} & 1.91M & 0.40M & 2.31M & 5,479 & \$0.53 \\
            \textbf{Microbiology} & 3.59M & 0.74M & 4.33M & 10,456 & \$0.98 \\
            \bottomrule
    \end{tabularx}
    \caption{Cost analysis for \ours using \texttt{gpt-4o-mini} for the Microbiology (most expensive) and US History subjects (least expensive).}
    \label{tab:cost-analysis}
\end{table*}

\section{\ourdata Data Processing Details}
\label{appendix:data_processing}

\subsection{Data License}
The openstax textbooks are licensed under the Creative Commons Attribution License. The textbooks used in this work are all in English. 

\subsection{Parsing Sections}
\label{appdx:parsing-sections}

Our prompt for parsing sections as part of creating \ourdata is shown below. 
In order to ensure consistency across subjects in how sections are divided, we provide manually annotated few-shot examples. 
In addition, we use GPT-4o for this part as it is a one-time expense and data quality has important implications for all the experiments that we conduct in this work.  

\begin{tcolorbox}[title=Extracting Sections, myboxstyle, breakable]
\begin{lstlisting}
Instructions for extracting sections from the given textbook content:

1. Transform markdown for equations into LaTeX and remove all other markdown
   formatting to only keep the raw content.
2. Split the content into sections of uniform length and number each section.
3. Skip the learning objectives, key concepts, and summary content.
4. Ensure that all content, except skipped parts, is covered verbatim in at
   least one of the resulting sections.

# EXAMPLE
## INPUT:
{{ example_input }}

## OUTPUT:
{{ example_output }}

Produce only valid JSON with the following format:
{"section": {"1": {"content": "Verbatim section 1 content from chapter"}, ...}}

# TARGET TEXTBOOK CONTENT:
{{ target_content }}
\end{lstlisting}
\end{tcolorbox}
\label{appdx:section-prompt}

\subsection{Parsing Exam Questions}
\label{appdx:parsing-exam-questions}

We parse the end-of-chapter review questions in OpenStax textbooks based on the markdown headers and formatting with BeautifulSoup.\footnote{\url{https://www.crummy.com/software/BeautifulSoup/}}
There are occasionally ill-formatted questions that we throw out, since the main goal here is to find a subset of chapters that are suitable for testing \ours. 
We avoid chapters with too few ($<10$) and too many questions ($>25$) so that our simulations are conducted with enough questions for measuring question utility while also completing within a reasonable timeframe.  

\subsection{Bloom's Taxonomy Level Distribution}
\label{appdx:bloom-taxonomy-level-distribution}
We share our prompt for categorizing questions based on the revised Bloom's taxonomy below: 
\begin{tcolorbox}[title=Bloom Classification, myboxstyle, breakable]
\texttt{
\noindent Classify the questions into one of the six main categories of Bloom's 
Taxonomy based on the cognitive processes required for answering it correctly.  
\\
\\
\textbf{Bloom's Taxonomy Categories:}  
\begin{enumerate}
    \item \textbf{Remembering:} Producing or retrieving definitions, facts, or 
          lists, or reciting previously learned information.  
    \item \textbf{Understanding:} Grasping the meaning of information by 
          interpreting and translating what has been learned.  
    \item \textbf{Applying:} Using learned information in new and concrete 
          situations.  
    \item \textbf{Analyzing:} Breaking down or distinguishing the parts of 
          learned information.  
    \item \textbf{Evaluating:} Making judgments about information, validity of 
          ideas, or quality of work based on a set of criteria.  
    \item \textbf{Creating:} Using information to generate new ideas or 
          products.  
\end{enumerate}
}
\begin{lstlisting}
Question 1: {{question 1}}
Question 2: {{question 2}}
...
\end{lstlisting}

\texttt{Provide only the Bloom category and format your response in JSON with the 
following structure:}

\begin{lstlisting}
{"bloom_categories": [
    {"question": question, "bloom_category": bloom_category}
]}
\end{lstlisting}

\end{tcolorbox}

\begin{figure}[t]
    \centering
    \includegraphics[width=\columnwidth]{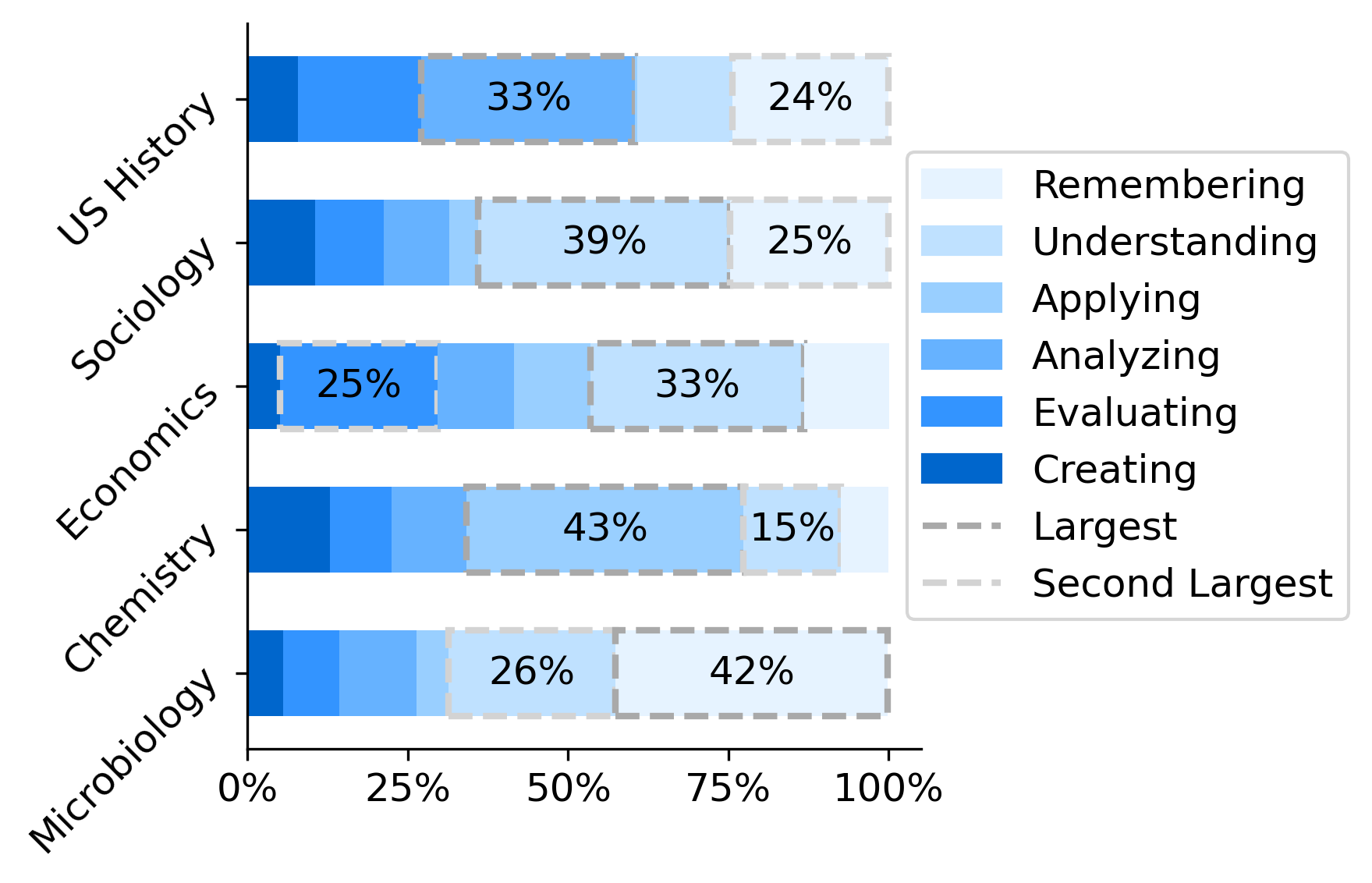}
    \caption{\textbf{Distribution of Bloom’s taxonomy labels in \ourdata.} \ourdata consists of questions that require a wide variety of cognitive levels and subjects vary in the cognitive skills emphasized in their exams.}
    \label{fig:bloom-distribution}
\end{figure}

With our prompt, each question is mapped to one of six categories: \textit{Remembering}, \textit{Understanding}, \textit{Applying}, \textit{Analyzing}, \textit{Evaluating}, or \textit{Creating}, and aligned to relevant document sections.
The distribution, shown in Figure \ref{fig:bloom-distribution}, indicates that \ourdata includes a broad range of question types, supporting outcome-based evaluation across different cognitive levels.
For instance, questions in Microbiology and Sociology primarily focus on \textit{Remembering} and \textit{Understanding}, whereas Chemistry and Economics exhibit a more varied distribution.


\section{Baseline Details}
\label{appdx:baseline-details}

In this section, we provide additional details for the baselines that we compare against \ours-trained models. 

\subsection{Bloom-based Prompting}
\label{appdx:bloom-based-prompting}

\begin{tcolorbox}[title=Generating Next Paragraph, myboxstyle]
\begin{lstlisting}
Use the cognitive process of {{selected_bloom_level}}: {{bloom_definition}}
to generate the next paragraph for the following
text that will help the student understand the content better:

{{full_context}}

Output in the following JSON format:
{"next_paragraph": next_paragraph}
\end{lstlisting}

\end{tcolorbox}

\begin{tcolorbox}[title=Generating Question, myboxstyle]
\begin{lstlisting}
Given the input context and the next paragraph,
what is the key question that connects the two?

Input context: {{context}}
Next paragraph: {{next_paragraph}}

Output in the following JSON format:
{"question": question}
\end{lstlisting}

\end{tcolorbox}

\section{Alternative Question Quality Metrics}
\label{appendix:evaluation-metrics}

\subsection{Salience}
\label{appendix:salience}
\textbf{Salience} measures a question’s relevance and importance within the document \( D \)~\cite{wu2024questions}.
It is rated on a Likert scale from 1 to 5, where a score of 1 indicates that the question in section \( k \) is unrelated to \( S_{[1:k]} \) and contributes minimally to understanding, while a score of 5 indicates strong relevance to \( S_{[1:k]} \) and essential comprehension support for \( S_k \) by clarifying key concepts or introducing new information.
We use the QSalience model released by \citet{wu2024questions}\footnote{\parbox[t]{\dimexpr\columnwidth-2.2em\relax}{\raggedright Available at \texttt{github.com/\allowbreak ritikamangla/\allowbreak QSalience}.}} to estimate salience. 
The model we use is a Mistral model fine-tuned with Peft\footnote{\parbox[t]{\dimexpr\columnwidth-2.2em\relax}{\raggedright\texttt{huggingface.co/\allowbreak lingchensanwen/\allowbreak mistral-ins-\allowbreak generation-\allowbreak best-balanced}.}}, which has a context window limit of 4096 tokens. 
Therefore, for inputs longer than the limit, we use \texttt{gpt-4o-mini} with the following prompt as a fallback:

\begin{tcolorbox}[title=Salience Prediction, myboxstyle, breakable]
\textbf{Article:} \textit{\{article\}}  \\
\textbf{Question:} \textit{\{question\}}  \\

\textbf{System Instructions}  
Imagine you are a curious reader going through the article. You come across a question and need to determine whether it should be answered within the article or not. Your task is to assign a score based on the relevance and necessity of answering the question. \\

\textbf{Scoring Criteria}  
\begin{itemize}
    \item \textbf{Score = 1:} The question is \textbf{completely unrelated} to the article.
    \item \textbf{Score = 2:} The question is \textbf{related but already answered} in the article.
    \item \textbf{Score = 3:} The question is \textbf{related but answering it is not essential}, as it expands on a \textbf{minor or non-central idea}.
    \item \textbf{Score = 4:} The question is \textbf{related and answering it enhances the reader’s understanding} of the article.
    \item \textbf{Score = 5:} The question is \textbf{related and must be answered}, as it expands on \textbf{central ideas of the article}.
\end{itemize}

\textbf{Scoring Guidelines}  
\begin{itemize}
    \item The score is based on the \textbf{information utility} of the answer.
    \item If a question is related but \textbf{not central or necessary}, \textbf{do NOT} assign it a high score.
    \item Assign \textbf{Score 3} if the question is unanswered but \textbf{not critical}, and \textbf{Score 2} if it has already been answered.
    \item \textbf{Distinguishing Scores 4 and 5:}
    \begin{itemize}
        \item If the article would feel \textbf{incomplete} without the answer, assign \textbf{Score 5}.
        \item Otherwise, assign \textbf{Score 4}.
    \end{itemize}
    \item A \textbf{Score of 4} is useful, but \textbf{other questions may be more important}.
    \item A \textbf{Score of 5} is reserved for \textbf{must-answer, central questions}.
    \item \textbf{Avoid bias toward high scores} and carefully follow the instructions.
\end{itemize}

The score should strictly be an \textbf{integer between 1 and 5}.

\textbf{Score:}

\end{tcolorbox}

\subsection{Expected Information Gain (EIG)}
\label{appendix:eig}
Expected Information Gain (EIG) quantifies the reduction in uncertainty about a student's knowledge state after answering a question.
We estimate EIG using an LLM by computing the entropy of the model’s token probability distribution before and after conditioning on the first token of the correct answer. EIG is computed as:
\[
EIG(Q) = H(X) - H(X | A_1)
\]
where \( H(X) \) is the prior entropy, representing the model’s uncertainty before seeing the answer, and \( H(X | A_1) \) is the posterior entropy after conditioning on the first token \( A_1 \). The entropy is given by:
\begin{align*}
H(X) &= - \sum_{x \in V} P(x) \log P(x), \\
H(X | A_1) &= - \sum_{x \in V} P(x | A_1) \log P(x | A_1)
\end{align*}
where \( V \) is the vocabulary, and \( P(x) \), \( P(x | A_1) \) are token probabilities before and after observing \( A_1 \), respectively.
We estimate probability distributions by querying an LLM with the following prompts.
Instead of conditioning on the full answer, we use only the first token, as most uncertainty reduction occurs at the start of an answer.
This aligns with the autoregressive nature of LLMs and ensures EIG captures incremental belief updates rather than full-answer memorization.

\begin{tcolorbox}[title=Prior Entropy Estimation, myboxstyle]
Imagine you are a reader encountering a question in the article.

\textbf{Article:} \{article\}

\textbf{Question:} \{question\}

\textbf{Answer:} 
\end{tcolorbox}

\begin{tcolorbox}[title=Posterior Entropy Estimation, myboxstyle]
Imagine you are a reader encountering a question in the article.

\textbf{Article:} \{article\}

\textbf{Question:} \{question\}

\textbf{Answer:} \{$A_1$\}
\end{tcolorbox}

\section{Additional Results}

\subsection{Qualitative Examples}
\label{appendix:qualitative-examples}

Table~\ref{tab:utility_analysis} extends the Chemistry excerpt in Table~\ref{tab:utility_examples_excerpt} with examples of questions that have high salience and high utility, high utility and low salience, and high salience and low utility across all five subjects. 
As we have shown with our results that compute a lower correlation between the two, there is no noticeable pattern based on a manual inspection.  
We encourage future work to refine question utility or develop new question quality metrics that provide more interpretable insight into the value of a question as it pertains to a specific downstream task.  

\begin{table*}[t]
    \centering
    \renewcommand{\arraystretch}{1.1}
    \resizebox{\textwidth}{!}{%
    \begin{tabular}{@{} >{\centering\arraybackslash}m{2.4cm} l @{}}
        \toprule
        Subject &
        \begin{tabular}{@{} >{\centering\arraybackslash}m{1.7cm} >{\centering\arraybackslash}m{1.5cm} p{0.58\textwidth} @{}}
            Salience & Utility & Question
        \end{tabular} \\
        \midrule
        Chemistry &
        \begin{tabular}{@{} >{\centering\arraybackslash}m{1.7cm} >{\centering\arraybackslash}m{1.5cm} p{0.58\textwidth} @{}}
            High & High & How does the combination of elements result in properties that are different from those of the individual elements in their uncombined states, and can you provide more examples of such transformations? \\[0.7em]
            Low & High & What are the `Twelve Principles of Green Chemistry'? \\[0.7em]
            High & Low & What happens to the charge of an atom when it gains or loses electrons?
        \end{tabular} \\
        \midrule
        Economics &
        \begin{tabular}{@{} >{\centering\arraybackslash}m{1.7cm} >{\centering\arraybackslash}m{1.5cm} p{0.58\textwidth} @{}}
            High & High & Why does inelastic demand allow the government to pass the tax burden onto consumers without significantly affecting the quantity demanded? \\[0.7em]
            Low & High & What are the three different combinations of labor and physical capital mentioned for cleaning up a park? \\[0.7em]
            High & Low & How does Jos\'{e} evaluate the trade-off between giving up the third T-shirt and purchasing two additional movies in terms of marginal utility, and what does this reveal about his preferences for consumption?
        \end{tabular} \\
        \midrule
        Sociology &
        \begin{tabular}{@{} >{\centering\arraybackslash}m{1.7cm} >{\centering\arraybackslash}m{1.5cm} p{0.58\textwidth} @{}}
            High & High & How does the concept of carrying capacity relate to the tragedy of the commons as described by Garrett Hardin and William Forster Lloyd? \\[0.7em]
            Low & High & What are some reasons people often forget where their old electronics are disposed of or stored? \\[0.7em]
            High & Low & How do the challenges faced by students from low socioeconomic backgrounds, as described in the example, impact their ability to succeed in the educational system compared to their peers from higher social classes?
        \end{tabular} \\
        \midrule
        U.S. History &
        \begin{tabular}{@{} >{\centering\arraybackslash}m{1.7cm} >{\centering\arraybackslash}m{1.5cm} p{0.58\textwidth} @{}}
            High & High & What were the main differences in the motivations and goals of the English migrants who settled in the Chesapeake Bay colonies compared to those who settled in New England? \\[0.7em]
            Low & High & What challenges did Venetian sailors face when transporting goods along the old Silk Road? \\[0.7em]
            High & Low & What were some of the challenges and dangers faced by serfs and their families in their daily lives during the Middle Ages?
        \end{tabular} \\
        \midrule
        Microbiology &
        \begin{tabular}{@{} >{\centering\arraybackslash}m{1.7cm} >{\centering\arraybackslash}m{1.5cm} p{0.58\textwidth} @{}}
            High & High & What roles do NAD+, NADP+, and FAD play in cellular metabolism, and how do their oxidized and reduced forms differ in function? \\[0.7em]
            Low & High & What evidence did Robert Remak provide to support the idea that cells originate from other cells? \\[0.7em]
            High & Low & What was the significance of Dmitri Ivanovski's use of the Chamberland filter in discovering the cause of tobacco mosaic disease?
        \end{tabular} \\
        \bottomrule
    \end{tabular}
    }
    \caption{\textbf{Qualitative examples} of generated questions by salience and utility. Each subject includes one high-salience/high-utility, one low-salience/high-utility, and one high-salience/low-utility question.}
    \label{tab:utility_analysis}
\end{table*}

\end{document}